\documentclass[sigconf]{aamas} 

\usepackage{balance} 
\usepackage[inline]{enumitem}

\newcommand{\wrt}{{\it w.r.t. }}    

\setcopyright{ifaamas}
\acmConference[AAMAS '21]{Proc.\@ of the 20th International Conference on Autonomous Agents and Multiagent Systems (AAMAS 2021)}{May 3--7, 2021}{Online}{U.~Endriss, A.~Now\'{e}, F.~Dignum, A.~Lomuscio (eds.)}
\copyrightyear{2021}
\acmYear{2021}
\acmDOI{}
\acmPrice{}
\acmISBN{}

\acmSubmissionID{587}

\title[Interrogating the Black Box]{Interrogating the Black Box: Transparency through Information-Seeking Dialogues}

\author{Andrea Aler Tubella}
\affiliation{
  \institution{Umeå University}
  \city{Umeå, Sweden}}
\email{andrea.aler@umu.se}

\author{Andreas Theodorou}
\affiliation{
 \institution{Umeå University}
  \city{Umeå, Sweden}}
\email{andreas.theodorou@umu.se}

\author{Juan Carlos Nieves}
\affiliation{
  \institution{Umeå University}
  \city{Umeå, Sweden}}
\email{juan.carlos.nieves@umu.se}

\begin{abstract}

This paper is preoccupied with the following question: given a (possibly opaque) learning system, how can we understand whether its behaviour adheres to governance constraints? The answer can be quite simple: we just need to ``ask'' the system about it. 
We propose to construct an \textit{investigator agent} to query a learning agent-- the \textit{suspect agent}-- to investigate its adherence to a given ethical policy in the context of an information-seeking dialogue, modeled in formal argumentation settings. This formal dialogue framework is the main contribution of this paper. Through it, we break down compliance checking mechanisms into three modular components, each of which can be tailored to various needs in a vast amount of ways: an investigator agent, a suspect agent, and an acceptance protocol determining whether the responses of the suspect agent comply with the policy. This acceptance protocol presents a fundamentally different approach to aggregation: rather than using quantitative methods to deal with the non-determinism of a learning system, we leverage the use of argumentation semantics to investigate the notion of properties holding $\textit{consistently}$.  Overall, we argue that the introduced formal dialogue framework opens many avenues both in the area of compliance checking and in the analysis of properties of opaque systems.

\end{abstract}


\begin{document}


\pagestyle{fancy}
\fancyhead{}


\maketitle

\section{Introduction}\label{S:introduction}

With the rise of the use of intelligent systems in all facets of public and private decision-making, methods to guarantee \textit{accountability, responsibility} and \textit{transparency} in the design and use of these systems are urgently needed \cite{Dignum2019}. These entail a duty to develop intelligent systems which respect fundamental human principles and values, are aligned with the social expectations of their deployment area, provide guarantees, and exhibit transparency. However, ethical values and principles are highly dependent on the socio-cultural context and their interpretations may differ for each stakeholder~\cite{Turiel2001TheMorality}. To ensure accountability and transparency, it is therefore fundamental that these interpretations are made explicit in the form of requirements \cite{Aldewereld2015DesignDevelopment}, and that we are able to audit and explain how the system follows them. This paper is thus preoccupied with the following question: given a (possibly opaque) learning system, how can we understand whether its behaviour adheres to such requirements? We argue that the answer can be quite simple: we just need to ``ask'' the system about it. 

In this paper, we propose a formal dialogue framework to investigate a learning system and evaluate its responses for compliance with a policy. We construct an \textit{investigator agent} and a \textit{suspect agent}, which together enact an information-seeking dialogue \cite{Walton1995CommitmentReasoning} describing the behaviour of the learning system (Figure \ref{fig:framework}). The extracted information about this behaviour is then modelled in formal argumentation settings and evaluated for compliance with the policy. The proposed dialogue framework is greatly versatile: it integrates three components-- an investigator agent, a suspect agent, and acceptance criteria-- each of which can be tailored to various needs in a vast amount of ways. In this way, we offer an inspection mechanism that can be adjusted by adapting each modular component, allowing us to tune all aspects of:
\begin{enumerate*}
    \item[i)] query generation,
    \item[ii)] interpretation of knowledge obtained from the system being inspected, and
    \item[iii)] acceptance criteria for compliance.
\end{enumerate*}

Our framework leverages the structure of information-seeking dialogues and of formal argumentation frameworks to address the challenge of transparent evaluation and audit of learning systems. In addition to the benefit of modularity, the use of this framework brings a distinct advantage in two areas: the transparency of the evaluation process itself and the handling of inconsistent (or non-deterministic) behaviour of learning systems. Indeed, to ensure accountability it is fundamental that evaluation and auditing procedures are transparent and explainable. By conducting this process as a dialogue, we are able to present an evaluation process that is directly human-accessible and transparent. Additionally, a specific challenge for both the explainability and auditability of learning systems is that of consistency, in the sense that the properties of the output are not necessarily consistent for inputs sharing the same features, and may even vary with time for identical inputs. Often, this variability is approached with quantitative statistical methods, describing desirable behaviour in terms of averages and distances. The approach we propose is complementary and fundamentally different, consisting on modelling the behaviour of a learning system in a formal argumentation framework. From a formal perspective, argumentation frameworks are well equipped to deal with inconsistency, and are often used to model human dialogues which are naturally inconsistent. By making use of them, we are able to investigate what it could mean for a policy to \textit{consistently} describe the behaviour of the learning system.

This paper is structured as follows. First, we discuss the necessary background. Then, we introduce a running example, that we use in the following section to illustrate the theory. Next, we exemplify the possibilities of the formal dialogue framework in an example implementation. Finally, we discuss related work and future research directions. Overall, we argue that the introduced formal dialogue framework opens many avenues both in the area of compliance checking, as well as in the analysis of properties of opaque systems.

\section{Motivational Background}\label{S:Background}
In this section, we outline the key challenges and literature we took into consideration for developing our dialogue framework. 

\subsection{Compliance checking in the responsible design of intelligent systems}
Simulators, testing procedures, and other safety measures can only cover \textit{what the developers thought of}. The emerging behaviour of an agent as its interacts with its environment is far more complex. Incidents, either due to misuse or malicious use, with autonomous systems are bound to happen. We should work at both minimising them and be able to assign---needed-- \textit{accountability} to the manufacturers and users of artefacts \cite{BrysonTheodorou2019}. To do so, we need to be able to audit our systems to understand how a system complies---or not---with our legal values and what went wrong. The auditability of our systems is often linked to having an adequate implementation of a \textit{transparency} \cite{BrysonWinfield2017}.

A different---but related---benefit of transparency is that it enables real-time calibration of trust between the human users and their autonomous systems \cite{Dzindolet2003,Sanders2014}. That is by providing additional information regarding the system, its users are able to create a more accurate mental model about the system and adjust their expectations accordingly \cite{Theodorou2017ConnectionScience}. Calibration of trust enables the user to adjust their expectations---or even to predict certain actions from the system---reducing misuse or disuse of the system \cite{Lee2004}.

In the literature, there are multiple approaches at providing---and defining---transparency at the time of operation. Examples include the communication of information regarding the machine's abilities \cite{Mercado2016} and capabilities \cite{Wohleber2017}, the system providing alerts to the user \cite{Kim2006}, or by providing information related for the features which are responsible for the prediction result \cite{Biran2017, Anjomshoae2019}. Others, suggest a post-incident approach, where we review information logged by our system for traceability purposes \cite{Winfield2017}.

While these approaches work, it may not be always possible---or desirable---to implement them \cite{Ananny2018}. The reasons might be technical (e.g. the algorithm is not easy to explain), economic (e.g. costs of developing a transparency mechanism for the specific model may be excessive), commercial (e.g. concerns over compromising of trade secrets), or social (e.g. revealing input may violate privacy expectations). Standards, such as the \textit{IEEE P7001 Standard for Transparency of Autonomous Systems}, recognise it and work around these issues by providing various levels of compliance depending on the receiver of the transparency-related information \cite{BrysonWinfield2017}. This `social' solution does not solve the technical challenges of implementing transparency for certain approaches or directly addresses the costs associated with the need of a new transparency method for nearly every different AI technique; e.g. data-driven approaches require fundamentally different solutions from argumentation-based approaches---which are, arguably, inherently transparent. 

Even if we are able to determine \textit{how to audit} our system, we need to be able to understand \textit{what to audit them for}. The increased pervasiveness of intelligent systems in all areas of public and private decision-making has led to a considerable push in the publication of guidelines and standards guiding the design process of reliable and \textit{trustworthy} intelligent systems that align not only with the law, but also with our social values. Several organisations have been producing---and reproducing---high-level `Ethical Guidelines' to promote high-level abstract values, e.g. fairness, privacy, and others \cite{Theodorou2020TowardsAI}. Creating universally-accepted definitions of ethical and social values is impossible due to their contextual nature \cite{Turiel2001TheMorality}. Yet, definitions are necessary to produce technical requirements. Hence, 
it is fundamental to make the interpretations of values concrete and explicit \cite{Theodorou2020TowardsAI}.

In this paper, we provide a novel framework to provide a technical solution to the challenges of transparency and auditability of values. We consider a policy containing concrete and explicit requirements of the expected behaviour of the system. Our approach focuses specifically on the inputs and outputs of the system, alongside with its data---and in particular the relevant metadata---to test these requirements. Our proposed dialogue framework can accommodate a breadth of requirements and is by itself explainable. By making our own framework explainable, we ensure the explicitness of our values and enable them to be audited and, therefore, verified.


\subsection{Information-seeking dialogues and formal argumentation}

Within the idea of information-exchange as a dialogue, a particularly fruitful outlook is the taxonomy introduced by Walton and Krabbe \cite{Walton1995CommitmentReasoning}. In their work, dialogues are seen as an exchange of information which fulfils a shared goal and where each party also has its own aims, and classified depending on those objectives. Through this lens, evaluating an intelligent system for compliance with a policy is an exchange with the goal of sharing information, in which the auditor has the goal to obtain the relevant information from the system, and the system has the goal to provide all the information as requested. This type of exchange can be found explicitly in the taxonomy under the name of an \textit{information-seeking dialogue}.
 
Information-seeking dialogues are defined as an exchange in which one participant aims to obtain information it does not have from another participant, who is believed to possess this information. Thus, information-seeking dialogues are asymmetric: one agent has more information than the other on the specific topic of interest. In fact, it is the only such asymmetric class in Walton and Krabbe's taxonomy. This type of dialogues heavily relies on one of the agents (which we will call \textit{investigator}) having a specific topic they seek information about. This is particularly apt for our purpose: the information we seek is directly related to the ethical policy we are investigating. In addition, by conducting the evaluation process with the structure of a dialogue, we are able to provide a level of transparency about the process, by presenting it in a directly human-readable form.
 
Argumentation frameworks have widely been studied as a tool to model and generate dialogues \cite{FanToni14,BlackH09,ParsonsWA03}, by naturally providing a representation for the structure of arguments. A specific advantage of formal argumentation is brought forward by its ability to represent inconsistencies, which arise often in human dialogue. In the context of the evaluation of a learning system, we leverage the properties of argumentation frameworks and argumentation semantics to formally represent and reasoning about the possible inconsistencies in the system's behaviour. For example, a property rarely holds true for every single input belonging to a class. By modelling this type of inconsistency as an attack between arguments, we propose an option of what it could mean for a policy to \textit{consistently} describe the behaviour of a learning system.

\section{Running example}\label{S:RE}
Unlike common belief, recommender systems, i.e. systems that suggest to each user content that is relevant to them (with or without personalisation), are not free of ethical concerns. A growing body of research is breaching the topic of the ethics of recommender systems \cite{Milano2020RecommenderChallenges}, addressing topics such as privacy, content filtering and autonomy. For the sake of brevity, in our running example and of its implementation we will focus on a simple policy for a movie recommender system. We picked this application domain for two reasons: 1) this is one of the most popular application domains for new practitioners due to the popularity of the MovieLens Dataset and the Netflix Grand Prize \cite{dataset, Koren_Bell_2011}; and 2) we want to showcase the possibilities of the framework through a simple, but representative, example.

Given a user and a movie as input, the recommender system outputs a list of 10 movies that the user is predicted to enjoy. The system is trained on the most popular dataset used for recommender systems, the MovieLens Dataset, which contains, for each user, a list of scores given to different movies \cite{dataset}. In addition, the dataset contains the title, summary, genre, budget, IMDb id number\footnote{IMDb is a popular movie database owned by Amazon (\url{www.imdb.com})}, keywords, and credits for each movie. 

For this recommender we propose a policy focused on the concept of data-asymmetry to better illustrate the use of our framework. Because of a myriad of factors, data is very rarely evenly distributed across features. In the case of the database of movies we consider, for example, independently produced action movies with a female director are underrepresented compared to other categories. Scarcity of data for certain features can have a direct effect on the prediction quality, as having few similar data points affects the generalisability of predictions. In the case of our example, we wish to set a policy that states that the quality of the output for the class of independently produced action movies with a female director should be high, no matter how underrepresented it is. In particular, an attribute of high quality output for the movie recommender system we are considering is given by variety, meaning that a recommendation is considered good if it features a wide variety of genres. For this reason, our policy will consist of a single norm: \textit{ independently produced action movies with a female director must produce recommendations with at least 10 different genres.} 

This toy example is of course merely an illustration. However, norms of the form "input of class $A$ must produce output with properties $P$" can be related to many concepts in the fairness literature. It thus is our aim to showcase how our framework evaluates such rules.

\section{Theoretical Framework}\label{S:TF}

In this section, we will formally introduce the formal dialogue network, defining each of its components. To illustrate the formal definitions, we will make use of the running example described in the previous section.

\subsection{Learning functions}

In terms of the learning system being inspected, we describe it in the simplest way possible without assuming any particular attributes.

\begin{definition}

We assume data is generated according to an underlying stochastic process $p :  X \rightarrow  Q$, where $X$ is a $d$-dimensional feature space.

A \textit{learning system} is given by a parametrised learning function $\hat{f}_D: X \rightarrow Y$ that is optimal for a given definition of optimality with respect to a data set $D=\{(\mathbf{x}_i, p(\mathbf{x}_i))\}_{i=1}^n$, where $\mathbf{x}_i := [x_i^1, \dots,x_i^d]^{\top} \in  X$.

\end{definition}






\begin{example}
The movie recommender system described in the running example is given by a function $\hat{f_D}: X \rightarrow Y$, where $X$ is composed of vectors $(u,m)$ with $u$ a feature vector representing a user and $m$ a feature vector representing a movie, and where $Y$ is a set of sets of movies.
\end{example}

\subsection{Extended definite logic program}

An important part of the proposed dialogue framework hinges on the representation of the policy that we are evaluating compliance with. We will represent it as \textit{rules} that input/output pairs should follow. Formally, it is described by an extended definite logic program, in which each clause is such a rule. Using this language allows us to simplify the concepts in the policy into propositional atoms, which we will later use to generate topics.

\begin{definition}
The language of a propositional logic has an alphabet consisting of
\begin{description}
\item[(i)] propositional symbols: $\bot, \top, p_0, p_1, ...$
\item[(ii)] connectives : $\vee , \wedge , \leftarrow , \lnot , \; not$
\item[(iii)] auxiliary symbols : ( , ).
\end{description}
where $\vee , \wedge , \leftarrow$ are 2-place connectives, $\lnot$ is a 1-place connective and $\bot$, $\top$ are 0-place connectives. Propositional symbols and negated propositional symbols of the
form $\lnot p_i$ ($i \geq 0$) stand for the indecomposable
propositions, which we call {\it atoms}, or {\it atomic
propositions}. Atoms negated by $\lnot$ will be called \emph{extended atoms}. When we refer to atoms, we refer to both non-extended and extended atoms.



An \textit{extended definite clause} $C$, is denoted by
$$a  \leftarrow  a_1, \dots, a_{n}$$
\noindent where $n \geq 0$, and $a$, $a_i, 0\leq i \leq n$ are atoms. When $n=0$ the clause is an abbreviation of $a \leftarrow \top$ such that $\top$ is the proposition symbol that always evaluate to true. Sometimes we denote an extended definite clause {\it C} by $ a \leftarrow \mathcal{ B}$, where  $\mathcal{ B}$ is the set $\{a_1, \dots, a_{n}\}$.

\emph{An extended definite logic program} $P$ is a finite set of extended definite clauses. We denote the set of atoms in the language of $P$ by $\mathcal{L}_P$. Conversely, the set of all extended definite programs with atoms from $\mathcal{L}$ is denoted by $Prog_{\mathcal{ L}} $.
\end{definition}


\begin{example}\label{ex:policy}
The policy we presented in Section \ref{S:RE} is composed of a single norms that we will represent as a clause. The norm states that queries whose input is an independent action movie with a female director should produce as output a list of movies with at least 10 different genres.

We therefore set the clause
\[ \mathtt{highVariety(x)} \leftarrow \mathcal{B}\]
with $ \mathcal{B}=\{\mathtt{woman(director(x))},  \mathtt{independent(type(x))}, \\ \mathtt{action(genre(x))}\}$ where $\mathtt{x}$ is a variable representing an input/output pair. 

$\mathtt{woman(director(x))}$ is a propositional variable evaluated as true if the director of the movie provided as input is a woman, $\mathtt{independent(type(x))}$ evaluates as true if the input movie is an independent production, and $\mathtt{action(genre(x))}$ evaluates as true if the genre of the input movie is action. Finally, $\mathtt{highVariety(x)}$ is true when the list of movies in the output contains at least 10 genres.
\end{example}

Note that extended logic programs are assumed to be quantified over variables. That is, in the example above, the clause is assumed to hold for all $\mathtt{x}$.

\subsection{Arguments}

The main idea behind the framework we propose is to obtain information about the behaviour of a learning system in the form of arguments. In particular, the arguments considered will have the form of a pair: they will consist of a specific input, together with propositional variables describing it and/or its output.





\begin{definition}[A black-box argument]\label{def:LearntArgument}
Let $\hat{f}_D:X \rightarrow Y$ be a learning system with associated data set $D=\{(\mathbf{x}_i, p(\mathbf{x}_i))\}_{i=1}^n$, and $\mathcal{L}$ be a set of propositional atoms. 
A black-box argument is a tuple of the form $\langle \mathbf{x}_i, c\rangle$ where $c \in \mathcal{L}$.
Given a black-box argument $\langle \mathbf{x}, c \rangle$, $\mathbf{x}$ is called the support of the argument and $c$ its conclusion. 

We denote by ${ \mathcal{A}}_\mathcal{L}^{\hat{f}}$ the set of all the black-box arguments that can be built from ${\hat{f}_D}$ and $\mathcal{L}$.
\end{definition}

A fundamental contribution of this framework is the novel way we approach  the aggregation of information from the inputs and outputs of the learning system: what could it mean for a system to consistently follow a policy? An option is to opt to define consistency in a quantitative way, perhaps by setting that consistently following a policy means following it in 90\% of cases, or in 80\% of cases across features.  In this work, we will consider a notion of consistency based on argumentation semantics. We hold that consistency should mean that similar inputs produce the same properties with respect to a trait.  This definition relaxes the definition of determinism-- where identical inputs should produce identical outputs. It puts the onus of consistency on the notion of \textit{similarity}. This shift is an asset in terms of generality: the definition of similarity can be adapted depending on the learning system and policy we wish to check.

\begin{definition}[Similarity map]
 Let ${ D} := \{(\mathbf{x}_i,p(\mathbf{x}_i))\}_{i=1}^n$, be the data of a learning function $\hat{f}_{D}$. We define the input dat set ${ D}_{\mathbf{x}} =  \{\mathbf{x}_i\}_{i=1}^n$ as the projection of the first component of the pairs in $D$.
 
 A similarity map is a map $similar: { D}_{\mathbf{x}} \times {D}_{\mathbf{x}} \rightarrow\{\top, \bot\}$. 
\end{definition}

\begin{example}
For our running example, we can implement a notion of similarity between movies based on the cosine distance accross keywords. We say that two inputs $(u_1, m_1)$ and $(u_2, m_2)$ are similar if $u_1=u_2$, and the distance between $m_1$ and $m_2$ is inferior to a threshold. This notion of similarity is particularly adapted to our norm: any input similar to an input from an unrepresented class with scarcity of data points will either belong to that class itself or suffer from the same issue of lack of similar data points. Thus, we wish to check that the norm that dictates that such inputs should produce varied outputs holds through this definition of similarity.

Of course, we could define similarity in a completely different way by, for example, dividing the input space into classes, and setting that two inputs are similar if they belong to the same class. In that case, through the formal dialogue framework we could assess if a property is held consistently across a class of inputs.
\end{example}


Intuitively, similar/input output pairs with conflicting properties with respect to the policy will be represented by arguments that attack each other, providing a representation of inconsistencies of the system with respect to the policy.

\begin{definition}[Conflicts between black-box arguments]\label{def:attack}

Let $Ar_1 = \langle \mathbf{x}_1, c_1 \rangle$ and $Ar_2 = \langle \mathbf{x}_2, c_2 \rangle$ be two arguments in a set of black box arguments ${ A}_d^{\hat{f}}$. We say that $Ar_1$ attacks $Ar_2$ if 
\[ similar(\mathbf{x}_1, \mathbf{x}_2) = \top \ \mathrm{and} \ c_1 \neq c_2 .\]

\end{definition}

This is a very simple notion of attack as we consider similar inputs whose descriptors of their output are not identical to be incompatible. Note that this automatically implies that the attack relationship is symmetrical: arguments attack each other. Although it is sufficient for our example, the attack relation can be tailored to each policy, and does not need symmetry. One could, for example, set the attack relation based on a semantic notion of which descriptors $c$ are incompatible with which others, rather than simply using the inequality relation. The definition of conflicting arguments is an important component in the versatility of the framework: it is where domain knowledge is encoded. In this sense, it can be made as sophisticated as desired.

In an information-seeking dialogue, the agent providing information is expected to provide arguments related to a specific topic. In our case, the topic will be given by a subset of acceptable inputs and a set of propositional atoms, which intuitively are the accepted descriptors of the input/output pair that fall within the topic.

\begin{definition}

Let $Topic=(T_X, T_P)$ where $T_X\subseteq X$ is a subset of \textit{acceptable inputs} of a $d$-dimensional space $X$, and $T_P\subseteq \mathcal{L}$ is a set of propositional atoms. We say that a black box argument $\langle \mathbf{x}, c \rangle$ is related to $Topic$ if $\mathbf{x}\in T_X$ and $c\in T_P$.

\end{definition}

\subsection{Information-seeking dialogues}

The framework we propose hinges on producing an information-seeking dialogue between an investigator agent and a suspect agent. To formally define the dialogue, we will adapt the move format introduced in \cite{BlackH09} to the context of our framework.

\begin{definition}
 We define the moves open, assert and close as the tuples described in the table below, where $a$ denotes an agent, $\langle \mathbf{x}, c \rangle$ is a black-box argument and $Topic= (T_X, T_P)$ where $F\subseteq X$ is a subset of \textit{acceptable inputs} of $X$, and $P$ is a set of propositional atoms.

\begin{center}
\begin{tabular}{ll} \\ \hline
Move & Format \\ \hline \hline
open & $\langle a, open, Topic \rangle$ \\
assert & $\langle a, assert, \langle \mathbf{x}, c \rangle \ \rangle$ \\
close & $\langle a, close \rangle$ \\ \hline
\end{tabular}
\end{center}

For each move instance $m$, we say that $a$ is its \textit{sender}, and denote it by $Sender(m)$.

\end{definition}

 Dialogues will be composed of moves in an ordered manner. An intitator agent will open a dialogue by setting a topic. Then, other agents will reply with arguments related to the topic. When an agent has no more arguments to bring forward, they will signal it by a close move.

\begin{definition}[Information-seeking dialogue]\label{def:dialogue}
A dialogue $\gamma$ is a tuple of the form $\langle { I}, D^t \rangle$ in which 
$D^t$ is an ordered sequence of moves $[m_1, \dots, m_t]$ involving a set of participating agents ${ I}$ such that $Sender(m_s) \in { I} \; (1 \leq s \leq t)$.

Let  $\gamma = \langle { I}, D^t \rangle$ be a dialogue. $\gamma$ is a \textit{well-formed information-seeking dialogue} if the following conditions hold true:

\begin{itemize}
  \item $m_1$ is an open move $\langle a, open, Topic \rangle$;
  \item $m_{2}, \dots m_{t- |{ I}|}$ are assert moves.
  \item If $m$ is an assert move in $D^t$,  then its black box argument is related to $Topic$, 
  \item $m_t,m_{t-1},\dots  ,m_{t-(|{ I}| -1)}$ are close moves.
  \item  $Sender(m_t) = Sender(m_1)$.
\end{itemize}

We say that $Topic$ is the topic of the dialogue.
\end{definition}

In the framework we present in this paper, we will consider information-seeking dialogues between two agents, an investigator and a suspect. The definition does however not constrain the number of participants, and it would be possible to add a third dialogue agent that for example has external information on the behaviour of the learning system.

Additionally, an interesting case of information-seeking dialogues is one where there is only one participant. Such a self-reflective dialogue can be used by an agent to ''query" itself.

\begin{definition}
 Let  $\gamma = \langle { I}, D^t_r \rangle$ be a well-formed information-seeking dialogue. $\gamma$ is a self-reflective dialogue if $|{I}| = 1$. 
\end{definition}

From the arguments shared in a dialogue, we will extract an argumentation graph that represents the characteristics of input/output pairs of the learning agent, and the inconsistencies encountered.

\begin{definition}
Let  $\gamma = \langle \mathcal{ I}, D^t \rangle$ be a dialogue and $A_{\gamma} = \{ \langle \mathbf{x}, c \rangle |$ $ \langle a, assert, \langle \mathbf{x}, c \rangle  \rangle \text { appears in }  D^t \}$. The argumentation graph related to $\gamma$ is the oriented graph $AF_{\gamma}= \langle A_{\gamma}, Att(A_{\gamma})\rangle$, where $Att(A_{\gamma})$ is composed of pairs $(Ar_1, Ar_2)$ where $Ar_1$ attacks $Ar_2$.
\end{definition}

Following Dung's style \cite{Dung95}, argumentation semantics are used for selecting arguments from an  argumentation graph $AF_{\gamma}$ related to a given dialogue $\gamma$. An argumentation semantics $\sigma$ is a function that assigns to an argumentation graph $AF_{\gamma}$ a set of sets of arguments denoted by $\mathcal{ E}_{\sigma} (AF_{\gamma})$. Each set of $\mathcal{ E}_{\sigma} (AF_{\gamma})$ is called $\sigma$-extension.
 $\sigma$ can be instantiated with any of the argumentation semantics that have been defined in terms of abstract arguments \cite{BaroniCG11}. 
 
 We can use the extracted semantics to understand whether the topic of questioning is relevant to the behaviour of the agent, i.e. if the propositional atoms used for the query can be consistently used to describe properties of input/output pairs (then we say the topic is \textit{sceptically accepted}). We can also relax this definition to consider that a topic produces relevant information if there is a description that can hold accross similar inputs (the topic is \textit{credulously accepted}). If the topic does not provide information about the learning system's behaviour because it does not hold across any similar inputs, we say that the topic is rejected.

\begin{definition}
Let  $\gamma = \langle \mathcal{ I}, D^t \rangle$ be a well-formed information-seeking dialogue with topic $Topic=(T_F, T_P)$. Let  $AF_{\gamma}$ be the argumentation graph related to $\gamma$ and $\sigma$ be an argumentation semantics.

\begin{itemize}
  \item $Topic$ is sceptically accepted \wrt $\sigma$ and $\gamma$ iff \\ $T_P \subseteq \bigcap_{E \in { E}_{\sigma} (AF_{\gamma})} \{c| \langle \mathtt{x}, c\rangle \in E\}$.
  \item $Topic$ is credulously accepted \wrt $\sigma$ and $\gamma$ iff \\ $T_P \subseteq \bigcup_{E \in \mathcal{ E}_{\sigma} (AF_{\gamma})} \{c| \langle \mathtt{x}, c\rangle \in E\}$.
  \item $Topic$ is rejected \wrt $\sigma$ and $\gamma$ iff  \\ $T_P \not\subseteq \bigcup_{E \in \mathcal{ E}_{\sigma} (AF_{\gamma})} \{c| \langle \mathtt{x}, c\rangle \in E\}$.
\end{itemize}
\end{definition}

Note, that the acceptance criteria we describe are fundamentally different than a quantitative approach studying how much of the dataset adheres to the policy.

\subsection{Formal dialogue framework }\label{S:FDF}

 \begin{figure}
    \centering
\includegraphics[scale=0.4]{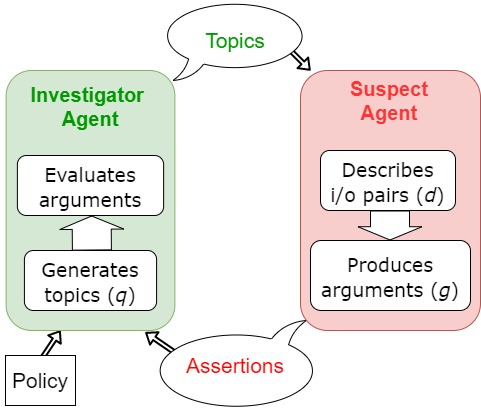}
    \caption{Formal dialogue framework. \\
    The tasks of the Investigator Agent and Suspect agent are described, with an arrow showing the order of occurrence. $g,d$ and $q$ denote concrete functions used by the agents, described in Section \ref{S:FDF}.}
    \label{fig:framework}
\end{figure}
 
The main contribution of this paper is a formal dialogue framework for the evaluation of learning agents, presented in Figure \ref{fig:framework}. Given a learning agent and a policy, the corresponding dialogue framework is composed of two agents: the \textit{investigator agent} and the \textit{suspect agent}. These two agents will engage in an information-seeking dialogue, for which the investigator agent will open a topic, and the suspect agent will respond with information it possesses related to this topic. We will show that we can use the information gathered through this dialogue to assess whether the policy sufficiently describes the behaviour of the system by applying formal argumentation methods to the arguments extracted from the dialogue.

The function of the investigator agent is to choose a topic for the information-seeking dialogue that will produce relevant information for  the policy that is being tested, i.e. query generation. In the setting of the dialogue, the role of this agent is thus to choose a \textit{topic}, to which all the arguments will have to relate to.

\begin{definition}[Investigator Agent]
An investigator agent is a tuple of the form $\langle P, \hat{f}_D, q \rangle$ where:
\begin{itemize}
  \item $P$ is an extended definite logic program denoting a policy,
  \item $\hat{f}_D:X \rightarrow Y$ is a learning system,
  \item $q$ is a map that to each clause $C$ in $P$ assigns a set of pairs $Topic=(T_X, T_P)$ such that $T_X\subseteq X$ is a subset of \textit{acceptable inputs} and $T_P\subseteq \mathcal{L}_P$ is a set of propositional atoms. $q$ is called a topic generator function.
\end{itemize}
\end{definition}

\begin{example}\label{ex:investigator}
An investigator agent for our running example is given by the policy $P$ containing a single clause $c$, the learning system $\hat{f}$, and a topic generator function $q(c)=\{ (X_w,V), (X_i, V),\\ (X_a, V), (X_w\cap X_i, V), (X_w \cap X_a, V), (X_i \cap X_a, V), (X_w \cap X_i \cap X_a , V) \}$,
where $X_w$ is the subset of inputs in which the input movie has a female director, $X_i$ is the subset of inputs in which the input movie is an independent production, $X_a$ is the subset of inputs in which the input movie is of the action genre, and $V=\{\mathtt{highVariety},\\ \mathtt{mediumVariety}, \mathtt{lowVariety}\}$.

This query generation function stems from an interest to understand how consistent properties are when inputs are made increasingly concrete, and thus get closer to an underrepresented class, a version of monotonicity described in Definition \ref{def:non-monotonic}. In this case, it is created from the policy by transforming propositions referring to the inputs into input spaces, and by choosing all the propositional atoms referring to variety (a semantical choice). 

Of course, the query generation function could be implemented in a variety of other ways: from using a game-theoretical approach to maximise the information obtained by the queries, to using a learning system that maximises query optimality. By leaving this query generation function general, we aim to explore the different possibilities it can provide.
\end{example}

Ideally, the agent answering queries would be the learning system itself, but unfortunately such agents do not often come with dialectical capabilities. Thus, we construct a suspect agent that encapsulates the learning agent and adds structure around it to turn it into a dialogue-enabled agent. A suspect agent therefore has two capabilities: it can translate input/output pairs into the language of propositional atoms (thus \textit{describing} the pairs) and it can produce arguments from these descriptions.

\begin{definition}[Suspect Agent]
A suspect Agent agent is a tuple of the form $\langle P, \hat{f}_{{ D}}, d, g \rangle$ where:
\begin{itemize}
    \item $P$ is an extended definite logic program denoting a policy,
  \item $\hat{f}_{{ D}}:X \rightarrow Y$ is a learning function,
  \item $d: Y \rightarrow  2^{\mathcal{L}_P}$ is a \textit{description map}, that for every output $Y$
returns a set of descriptors i.e. a set of propositional atoms in the language of the policy,
 \item $g$ is a map such that for every topic $Topic=(T_X, T_P)$ returns the set of all black box arguments $\langle \mathbf{x}, c \rangle$ with $c \in d \circ \hat{f}(\mathbf{x})$ related to $Topic$. $g$ is called an \textit{argument generator}.
\end{itemize}

\end{definition}

\begin{example}\label{ex:suspect}
A suspect agent for our running example is given by the policy, the learning system $\hat{f}$, the argument generator $g$, and a description map $d$ that returns $\texttt{highVariety}$ if the output contains more than 10 genres, $\texttt{mediumVariety}$ if the output contains 6 to 10 genres, and $\texttt{lowVariety}$ if the output contains 5 or less genres.
\end{example}


The description map can be expanded to encompass all types of descriptions. For example, it can describe features of the input, features of the output, or properties of the relationship between input and output. During the dialogue, only those descriptors that are part of the topic will be used in arguments.

Once the suspect agent has argued about a given topic $T$, the investigator agent can reason about the''mental states" of the suspect agent regarding $T$. To evaluate these ''mental states" of the suspect agent, the investigator agent will consider an argumentation semantics to evaluate the argument graph that was constructed by the  suggested arguments of the suspect agent.

\begin{definition}[Belief-checking]\label{def:belief-checking}
Let $Ag^S = \langle P, \hat{f}_{\mathcal{ D}}, d, g \rangle$ be a suspect agent, $Ag^I = \langle P, \hat{f}_D, q \rangle$ be an investigator agent and  $\sigma$ be an argumentation semantics. 

\begin{itemize}
  \item $Ag^S$ $\sigma$-sceptically argues about $Topic \in q(P)$ if there exists a well-formed information-seeking dialogue $\gamma = \langle \mathcal{ I}, D^t \rangle$ such that  $m_1= \langle Ag^I, open, Topic \rangle$, $\mathcal{ I} := \{Ag^I,Ag^S\}$ and $Topic$ is sceptically accepted \wrt $\sigma$ and $\gamma$.
  \item $Ag^S$ $\sigma$-credulously argues about $Topic \in q(P)$ if there exists a well-formed information-seeking dialogue $\gamma = \langle \mathcal{ I}, D^t \rangle$ such that $m_1= \langle Ag^I, open, Topic \rangle$, $\mathcal{ I} := \{Ag^I,Ag^S\}$ and $Topic$ is credulously accepted \wrt $\sigma$ and $\gamma$.
  \item $Ag^S$ $\sigma$-empty argues about $Topic \in q(P)$ if there is a well-formed information-seeking dialogue $\gamma = \langle \mathcal{ I}, D^t \rangle$ such that $m_1= \langle Ag^I, open, Topic \rangle$, $\mathcal{ I} := \{Ag^I,Ag^S\}$ and $Topic$ is rejected \wrt $\sigma$ and $\gamma$.
\end{itemize}

\end{definition}

Let us observe that there is a wide variety of argumentation semantics in the state of the art of formal argumentation \cite{BaroniCG11} that can be used in Definition \ref{def:belief-checking}. 

A fundamental property that can be verified in our dialogue-based approach is the non-monotonicity of a learning system. This non-monotonicity property can be verified by increasing that information that is provided to a suspect agent as a topic in a dialogue.

\begin{definition}[Non-monotonic Belief-checking] \label{def:non-monotonic}
Let $Ag^I =\\ \langle P, \hat{f}_D, q \rangle$ be an investigator agent, $Ag^S = \langle P, \hat{f}_{\mathcal{ D}}, d, g \rangle$ be a suspect agent,  and  $\sigma$ be an argumentation semantics.  $Ag^S$ is non-monotonic if there exist $Topic^1_P$ and $Topic^2_P$ in $q(P)$ such that $Topic^1_P \subset Topic^2_P$, $Ag^S$ $\sigma$-$X_1$ argues about $Topic_1$, $Ag^S$ $\sigma$-$X_2$ argues about $Topic_2$ and $X_1 \neq X_2$.
\end{definition}

Once an investigator agent has identified the mental states of a suspect agent regarding a given topic $P$, an investigator agent can verify consistency between a policy and the mental states of a suspect agent. The methods for determining whether the suspect agent is consistent with the policy effectively provide an acceptance policy: if the suspect agent's assertions are consistent with the policy for a given definition of consistency, then it is determined that the agent satisfactorily adheres to the policy.

\begin{definition}[Interrogation]\label{def:interrogation}
Let $Ag^S = \langle P, \hat{f}_{\mathcal{ D}}, d, g \rangle$ be a suspect agent, $Ag^I = \langle P, \hat{f}_D, q \rangle$ be an investigator agent and  $\sigma$ be an argumentation semantics.  

\begin{itemize}
  \item $Ag^I$ strongly believes in  $Ag^S$ if for all $Topic \in q(P)$, $Ag^S$ $\sigma$-sceptically argues about $Topic$ and $P \cup Topic_P \nvdash \bot$. 
  \item $Ag^I$ credulously believes in  $Ag^S$ if for all $Topic \in q(P)$, $Ag^S$ $\sigma$-credulously argues about $Topic$ and $P \cup Topic_P \nvdash \bot$. 
  \item $Ag^I$ strongly does not believe in  $Ag^S$ if for all $Topic \in q(P)$, $Ag^S$ $\sigma$-empty argues about $Topic$ and $P \cup Topic_P \nvdash \bot$. 
\end{itemize}  

\end{definition}

We can choose to determine that policy is being followed when $Ag^I$ strongly believes in  $Ag^S$, or when $Ag^I$ credulously believes in  $Ag^S$, depending on how strict we wish to be. These acceptance protocols are by no means exclusive: acceptance can be made to depend on other aspects of the argumentation semantics, or even other aspects of the dialogue itself.

The framework we have presented offers many possibilities due to the modularity of its components. We can tune several aspects of \begin{enumerate*}
    \item[i)] query generation,
    \item[ii)] aggregation of knowledge obtained from the system being inspected, and
    \item[iii)] acceptance criteria for compliance.
\end{enumerate*}
In addition, the output of this process is a dialogue, containing topics, assertions about this topic, as well as arguments extracted from this dialogue and whether these arguments are accepted. All of this output is directly human-readable, and provides a level of transparency about both the functioning of the learning system being inspected and the compliance checking process itself.

\section{Example implementation}\label{S:Ex}

In this section, we describe an implementation of the framework for a small subset of the dataset of the running example described in Section \ref{S:RE}. Our aim is to study the adherence of this recommender system to the policy $P$ described in Example \ref{ex:policy}. Our recommender system is trained on the whole dataset, but for brevity of exposition we limit our queries to a small subset of it. We need to emphasise to the reader that our contribution is not the recommender system, but rather the framework used to evaluate it.

The recommender system combines two popular approaches, \textit{content based} \cite{Aggarwal_2016} and \textit{ collaborative Filtering} \cite{Breese1998}, into a hybrid system that first finds similar movies as the one inputted by the user and then ranks them based on the users' profile. For the movie search, we use movies' metadata such as cast, crew, genre, and keywords to calculate the cosine similarity between movies. Once calculated, we take the 20 closest movies and then rank them using our collaborative filtering approach. Collaborative Filtering is based on the notion that users similar to other users would rate items the same way. We use Singular Value Decomposition (SVD) algorithm to create our model. SVD has been made popular since its use by the winning team of the Netflix Grand Prize winner \cite{Koren2009TheBS,Koren_Bell_2011}. Our SVD model predicts the ratings a user would give to the 20 movies selected by our content-based part of the system. The system then ranks them based on those predictions and present only the top 10 to the user. We selected these techniques mentioned below due to their robustness, speed, and commonality in movie recommender systems.

The formal dialogue framework for this recommender system is given by the investigator agent and suspect agent described in Examples \ref{ex:investigator} and \ref{ex:suspect}. We aim to study whether the investigator agent strongly/credulously believes in the suspect agent with relationship to the policy $P$. This depends on the existence of a well-formed dialogue for every $Topic \in q(P)$. As described in Example \ref{ex:investigator}, there are seven topics produced by the investigator agent. For each of these topics, such a dialogue is produced by having the suspect agent assert all of the black box arguments related to the topic as produced by the argument generator $g$. For example, for the topic $(X_w , V)$, the produced dialogue is shown in Table \ref{fig:MoveTable}.

\begin{table}[t]
    \caption{Dialogue between the Investigator Agent and Suspect Agent for the topic $(X_w , V)$.}
        \label{fig:MoveTable}
\begin{tabular}{ll} \\ \hline
Name & Move \\ \hline \hline
$m_1$ & $\langle Ag^I, open, (X_w , V) \rangle$ \\
$m_2$ & $\langle Ag^S, assert, \langle \mathbf{x_1}, \mathtt{highVariety} \rangle \ \rangle$ \\
$m_3$ & $\langle Ag^S, assert, \langle \mathbf{x_2}, \mathtt{highVariety} \rangle \ \rangle$ \\
$m_4$ & $\langle Ag^S, assert, \langle \mathbf{x_3}, \mathtt{mediumVariety} \rangle \ \rangle$ \\
& \vdots\\
$m_{12}$ & $\langle Ag^S, close \rangle$ \\
$m_{13}$ & $\langle Ag^I, close \rangle$ \\\hline
\end{tabular}
\end{table}

From this dialogue $\gamma$, we extract the arguments and their attack relations. In this case, arguments attack each other when their support is similar (same user, similar movie), but the descriptor is different. We therefore obtain an argumentation graph $AF_{\gamma}= \langle A_{\gamma}, Att(A_{\gamma})\rangle$ where:
\begin{itemize}
\item $A_\gamma =\{ 1, \dots , 10 \}$
  \item $Att(A_{\gamma}) = \{
  (2,8),		(8,2),
(2,9),		(9,2),
(2,6),		(6,2),
(2,3),		(3,2),$ \\
$(3,6),		(6,3),
(3,8),		(8,3),
(3,2),		(2,3),
(4,6),		(6,4),
(6,3),		(3,6),$ \\
$(6,8),		(8,6),
(6,4),		(4,6),
(6,2),		(2,6),
(8,6),		(6,8),
(8,3),		(3,8),$ \\
$(8,10),		(10,8),
(9,2),		(2,9),
(10, 8), (8,10) \} $
\end{itemize} 

We are denoting arguments by the number of the move on which the argument was presented by the suspect agent. Let us apply two classical argumentation semantics the so-called grounded and stable semantics \cite{Dung95} to $AF_{\gamma}$\footnote{We used the argumentation solver: \url{http://gerd.dbai.tuwien.ac.at/}.}: the results are shown in Table \ref{fig:extensions}.

\begin{table}[t] 
  \caption{Extensions for the argumentation graph $AF_\gamma$ for grounded and stable semantics.}
    \label{fig:extensions}
\begin{tabular}{|l|l|}
  \hline
  Argumentation  & Extensions\\ 
    semantics  & \\ \hline
 $\sigma_{ground} (AF_{\gamma})$  & $\{1,5,7\}$ \\ \hline
 $\sigma_{stable}(AF_{\gamma})$  & $\{9,8,7,5,4,1\}$  \\
  & $\{10,9,7,6,5,1\}$   \\
   & $\{10,7,5,4,2,1\}$  \\
     & $\{10,9,7,5,4,3,1\}$   \\ \hline
  \hline
\end{tabular}
\end{table}

From these results, we can observe that the recommender system  sceptically argues about the arguments $\{1,5,7\}$. Hence, the recommender system has strong beliefs on arguments such as $1:= \langle \mathbf{x_1}, \mathtt{highVariety} \rangle$. However, there are arguments such as $2 := \langle \mathbf{x_2}, \mathtt{highVariety} \rangle$ that is low represented in the stable extensions. Hence, the investigator agent can believe that the recommender system has low evidence about Argument 2. By using Definition \ref{def:interrogation} and the results of $\sigma_{ground} (AF_{\gamma})$ and $\sigma_{stable}(AF_{\gamma})$, the investigator agent can verify the compliance of different policies. 
Let us observe that the grounded and stable semantics are only two argumentation semantics from a big variety of argumentation semantics that exits in the state of the art of formal argumentation reasoning \cite{BaroniCG11}. Hence, the selection of a proper argumentation semantics for implementing an investigator agent can be a question on its own.

\section{Related work}\label{S:RW}
The perspective of testing whether a learning agent complies with a policy in fact sets this work within the general area of \textit{conformance testing}. Conformance testing approaches for ''black box" and adaptive systems are still being developed: a specific challenge is that of the breadth of the test space \cite{Constant2007IntegratingSystems}. The framework proposed in this paper is related to a breadth of literature on agents testing other agents, particularly those approaches which propose to construct an agent or a multi-agent system with the explicit purpose of testing another agent. For example, \cite{Nguyen2008ExperimentalSystems} propose to construct an agent that can generate tests from  the ontologies describing a MAS being tested, after which responses to these tests are verified. Similar in outlook, \cite{Silveira2013RationalAgents} propose a framework consisting of a multi-agent system made-up of a testing agent, a monitoring agent and agents representing the task environment, particularly focused on identifying goals that are not being met by the agent being tested.
 
In the sense that our framework produces an argumentation graph modelling the behaviour of the learning agent being inspected, our approach is also related to work on \textit{agents modelling other agents}. The literature is vast in  this topic, in the context of multi-agent systems especially, given that in collaborative or competitive scenario it is often needed to produce a model of other agents to predict what their behaviour will be \cite{Suryadi1999LearningDiagrams, Carmel1996OpponentSystems}. Formal argumentation methods are more often used to model communication between agents \cite{Reed2005TowardsCommunication} and agent knowledge \cite{Bentahar2010ARepresentation}, but there are approaches that use argumentation frameworks to build an \textit{opponent model} representing what another agent believes based on a dialogue \cite{Thimm2014StrategicSystems,Rienstra2013OpponentArgumentation,HadjinikolisOpponentDialogues}. These are similar in outlook to the framework we present: in a way, we are representing the learning agent's beliefs in the form of properties that hold for its input/output pairs, in what resembles building a machine theory of mind \cite{Rabinowitz2018MachineMind}. 

\section{Conclusions and future work}\label{S:FW}
In this paper we present a modular framework for evaluating a learning system's adherence to a policy. The formal dialogue framework we present is based on the idea of building an argumentation framework representing the arguments expressed in a dialogue between an investigator agent and a suspect agent. In this way, we construct a model of the learning agent by considering its properties across inputs. A strength of this approach is given by the modularity of its components, each of which can be implemented in a variety of ways depending on which properties of the learning system we wish to study. Additionally, the use of a dialogue as an information-seeking tool provides a level of transparency about the querying and testing process. Finally, we propose acceptance criteria determining adherence to a policy that are fundamentally different from the quantitative approaches often used: acceptance is determined through argumentation semantics, suggesting a new notion of "consistent compliance" to a policy. The limits of this approach lay where access to the learning system is limited: when it is not possible to construct inputs matching the policy or to describe outputs with the predicates given by the policy.

Future work is planned on several directions, exploiting both the versatility of the framework and the potential for studying specific logical properties of learning agents. An important development is to extend this to more sophisticated representations of policies, with more sophisticated languages allowing for capturing the complexity of social and ethical norms. Beyond that, a refinement of this framework would be to study the possibilities of topic generation and output aggregation. For example, a possibility is to exploit the topic generator function of the investigator agent to adversarially generate those topics that are expected to yield more inconsistencies in the dialogue. Another is to implement the description function of the suspect agents (describing inputs/outputs in terms of the policy) as a learning system itself, which learns which are good or bad outputs in terms of the policy. An exciting possibility is to implement a description function that aggregates several inputs, returning descriptions of the output together with a probability weight: this would provide a hybrid quantitative/qualitative approach to aggregating knowledge about a learning system.

A further interesting research avenue is to study the presentation of the ''degree of agreement" to a policy, in a way that is most useful to foster trust and promote transparency. In this paper we have proposed two possible degrees-- sceptical and credulous-- but many other possibilities exist. Additionally, we aim to exploit methods developed for this framework to study technical properties of learning agents, such as monotonicity or rationality, similarly to how we defined the notion of non-monotonic belief checking.

Overall, we believe this framework offers distinct benefits in terms of modularity and transparency, as well as opening the door to new non-quantitative ideas of compliance. Furthermore, it opens many possibilities in terms of future development for the purpose of better understanding, and controlling, the learning systems that are becoming increasingly pervasive in our society.



\begin{acks}
A. Aler Tubella was supported by the Wallenberg AI, Autonomous Systems and Software Program (WASP) funded by the Knut and Alice Wallenberg Foundation. Theodorou A. is funded by the Knut and Alice Wallenberg Foundation, grant agreement 2020.0221.

\end{acks}


 \bibliographystyle{ACM-Reference-Format} 
\bibliography{referencesAndrea,papers_jcns,ath}

\end{document}